\documentclass[lettersize,journal]{IEEEtran}
\usepackage{amsmath,amsfonts}
\usepackage{algorithmic}
\usepackage{algorithm}
\usepackage{array}
\usepackage[caption=false,font=normalsize,labelfont=sf,textfont=sf]{subfig}
\usepackage{textcomp}
\usepackage{stfloats}
\usepackage{url}
\usepackage{verbatim}
\usepackage{graphicx}
\usepackage{cite}
\usepackage{multirow}
\usepackage{graphicx}
\usepackage{booktabs}
\usepackage{enumitem}
\usepackage{colortbl}
\usepackage{adjustbox}

\usepackage{multirow}
\usepackage[table,xcdraw]{xcolor}
\usepackage[colorlinks, citecolor=blue, linkcolor=red, citecolor=blue, urlcolor=cyan]{hyperref}
\usepackage[capitalise]{cleveref}
\begin{document}

\title{AWM-Fuse: Multi-Modality Image Fusion for Adverse Weather via Global and Local Text Perception}

\author{Xilai Li, Huichun Liu, Xiaosong Li, Tao Ye, Zhenyu Kuang, Huafeng Li
\thanks{This research was supported by the Basic and Applied Basic Research of Guangdong Province (No. 2023A1515140077), the Natural Science Foundation of Guangdong Province (No. 2024A1515011880),the National Natural Science Foundation of China(No. 52374166),  and the Research Fund of Guangdong-HongKong-Macao Joint Laboratory for Intelligent Micro-Nano Optoelectronic Technology (No. 2020B1212030010). (Xilai Li and Huichun Liu contributions equally). (Corresponding author: Xiaosong Li.)}
\thanks{Xilai Li, Huichun Liu, Xiaosong Li and Zhenyu Kuang are with the Guangdong-HongKong-Macao Joint Laboratory for Intelligent MicroNano Optoelectronic Technology, Foshan University, Foshan 528225, China. (emails: 20210300236@stu.fosu.edu.cn; Feecuin@163.com; lixiaosong@buaa.edu.cn; kmustkzy@126.com). } 
\thanks{Tao Ye is with the School of Mechanical Electronic and Information Engineering, China University of Mining and Technology, Beijing 100083, China. (e-mail: ayetao198715@163.com).}
\thanks{Huafeng Li is with the School of Information Engineering and Automation, Kunming University of Science and Technology, China. (e-mail: hfchina99@163.com).} 

}

\markboth{Journal of \LaTeX\ Class Files,~Vol.~14, No.~8, August~2021}%
{Shell \MakeLowercase{\textit{et al.}}: A Sample Article Using IEEEtran.cls for IEEE Journals}


\maketitle

\begin{abstract}
Multi-modality image fusion (MMIF) in adverse weather aims to address the loss of visual information caused by weather-related degradations, providing clearer scene representations. Although less studies have attempted to incorporate textual information to improve semantic perception, they often lack effective categorization and thorough analysis of textual content. In response, we propose AWM-Fuse, a novel fusion method for adverse weather conditions, designed to handle multiple degradations through global and local text perception within a unified, shared weight architecture. In particular, a global feature perception module leverages BLIP-produced captions to extract overall scene features and identify primary degradation types, thus promoting generalization across various adverse weather conditions. Complementing this, the local module employs detailed scene descriptions produced by ChatGPT to concentrate on specific degradation effects through concrete textual cues, thereby capturing finer details. Furthermore, textual descriptions are used to constrain the generation of fusion images, effectively steering the network learning process toward better alignment with real semantic labels, thereby promoting the learning of more meaningful visual features. Extensive experiments demonstrate that AWM-Fuse outperforms current state-of-the-art methods in complex weather conditions and downstream tasks. Our code is available at \href{https://github.com/Feecuin/AWM-Fuse}{https://github.com/Feecuin/AWM-Fuse}.

\end{abstract}


\section{Introduction}
\label{sec:intro}
\IEEEPARstart{I}{nfrared} and visible image fusion (IVIF) aims to generate more expressive image representations utilizing information from multiple sources \cite{r3,r65,r77,r24,r107,r110}. This technique has found widespread application in downstream tasks, such as object detection \cite{r6} and semantic segmentation \cite{r9}.

Recent advances in Vision-Language Model (VLM) have demonstrated the effectiveness of guiding visual modeling with textual semantic information \cite{r80,r65,r81,r82,r83,r19,r104}. Text provides a structured semantic framework, offering precise descriptions that reveal crucial information about images. Incorporating textual descriptions into visual model development enables the model to associate objects with relevant information such as states within the scene, thereby improving its capacity to capture and interpret image content \cite{r84}. This strategy has also been introduced to the image fusion task \cite{r65,r80}, where text-based methods have shown improved multi-modality feature interaction and facilitated user interaction. Despite efforts to integrate VLM into image fusion and restoration for improved real-world performance, several challenges remain:

\begin{enumerate}[label=\arabic*)]
\item Limited textual embedding design: Existing studies \cite{r65,r85} use simple textual cues for degradation embedding, without fully exploring the potential of VLM. Complex scenes exhibit both global degradation and nuanced local variations in object color, motion, and blur, which simplified descriptions (e.g., captions) often fail to capture.                                                                                                                                       
\item Overemphasis on local details: In Response, Subsequent efforts \cite{r80,r93} generate detailed scene descriptions from multiple image labels, but this offten leads to overfitting to fine-grained details, reducing the adaptability to global patterns and limiting the generalizability of detailed image-specific descriptions to unseen conditions.

\item Insufficient robustness under adverse weather: Even with current VLM-based multimodality image fusion models \cite{r65,r80} struggle in adverse weather, where severe conditions introduce multiple degradations that affect visible images and cause loss of contrast in infrared images. Moreover, maintaining performance in adverse weather with unified weight remains challenging.
\end{enumerate}

To address these limitations, we propose AWM-Fuse, a high-fidelity multi-modality image fusion algorithm for adverse weather, based on global and local text perception.
Firstly, we enhance the scene understanding and detail-capturing ability of the model under severe weather conditions by incorporating global and local text-aware modules. The global module uses BLIP \cite{r86} to generate an image caption summarizing the overall scene, primary objects, and degradation (e.g., rain, haze, snow), providing a broad environmental context. Complementing this, the local text-aware module employs ChatGPT \cite{r87} to generate detailed descriptions of object features, motion states, and localized degradations. This combined approach facilitates more precise object identification and fusion in adverse weather conditions. 
For caption information, we utilize the CLIP text encoder for feature extraction \cite{r90}. While CLIP captures global semantics, BLIP additionally captures sequential information within the text, enabling finer-grained semantic alignment. 
The two-dimensional features extracted by CLIP represent the overall sentence structure and primarily capture global semantics, which makes it challenging to detect finer details. Therefore, we employ the text encoder of BLIP to extract features from the detailed text generated by ChatGPT.

Existing fusion methods \cite{r4,r12,r17,r24,r32,r88,r89} often prioritize visual modality and lack semantic alignment constraints. In contrast, CLIP, a powerful VLM pre-trained on large-scale image-text data, effectively captures image-text alignment. Leveraging this capability, we propose a novel VLM-driven loss function that utilizes the image-text matching of CLIP to align the fusion output with the textual description of a clean multi-modality image during training.  This image-text feature matching learning strategy enhances the robustness and adaptability of fused images across diverse weather conditions and degradation scenarios, improving real-world applicability.

\begin{itemize}
\item We proposed AWM-Fuse, a novel multi-modality image fusion method for adverse weather. AWM-Fuse leverages global and local text perception within a unified, shared-weight architecture to effectively handle multiple degradations simultaneously. 

\item We proposed novel global and local text perception modules that guide feature extraction from both macro and micro perspectives, enhancing generalizability while preserving high-fidelity fusion.

\item We propose a novel VLM-driven loss using CLIP to guide fusion image restoration. For the first time, we introduce rich and high-quality textual descriptions for multi-modality images under adverse weather.

\item We conduct extensive experiments under diverse adverse weather conditions, showcasing the state-of-the-art performance of AWM-Fuse and its practical applicability in real-world scenarios through various downstream tasks.
\end{itemize}

\section{Related Work}
\subsection{Multi-Modality Image Fusion}
In recent years, researchers \cite{r18,r22,r23,r24,r32,r56,r57,r64,r73} have focused on harnessing the synergy between infrared and visible modalities in IVIF.  However, because infrared and visible data represent cross-modal data, simple feature fusion can lead to pixel redundancy or over-reliance on one modality, diverging from human visual perception. To address this, decoupling shared and modality-specific features has become a widely adopted strategy. For example, CDDFuse \cite{r24} combines the global attention of Transformer with the unique local context extraction capabilities of CNN. Recently, researchers \cite{r88} have also explored higher-order modeling for IVIF tasks or have revisited the fusion generation process from an equal variance perspective \cite{r77}. Collectively, these methods provide a foundation for facilitating comprehensive multi-modality information interaction.

To enhance the real-world applicability of IVIF, many studies have adapted it for downstream tasks \cite{r79,r17,r5,r21,r109}. For example, Liu et al. \cite{r17} developed a two-layer optimization approach to jointly address fusion and target detection, enabling the preservation of salient target information in fusion results. Tang et al. \cite{r21} introduced a semantically-aware real-time fusion network, combining fusion with semantic segmentation to enrich fused images with semantic details. 
Additionally, deploying IVIF algorithms \cite{r91,r65,r27,r92,r108} in complex scenes has garnered significant attention. For instance, Li et al. \cite{r91} explored a unified framework that cascades image restoration and fusion tasks, and introduced the first IVIF architecture specifically tailored for adverse weather conditions. To cope with the effects of real-world noise, Huang et al. \cite{r92} proposed an image decomposition-driven network to jointly perform fusion and denoising tasks.

\subsection{Vision-Language Model for Low-level Vision Tasks.}
With the rapid advancement of deep learning, VLM are becoming increasingly relevant to low-level vision tasks, showcasing strong performance across various applications \cite{r93,r94,r80,r81,r82,r83}. Representative VLMs such as CLIP, BLIP, and ChatGPT-4 \cite{r87} offer complementary strengths: CLIP aligns visual and textual representations, enabling cross-modal retrieval and semantic interpretation of images; BLIP not only generates descriptive captions but also performs visual question answering, thereby enhancing scene-level understanding, and ChatGPT-4, as an advanced large language model, strengthens contextual reasoning and natural language generation. Together, these models highlight the potential of VLMs to bridge vision and language, expanding the possibilities for vision tasks guided by linguistic cues.
For example, Zhao et al. \cite{r80} proposed a VLM-based image fusion framework that guides the fusion process using text information in the source image. Luo et al. \cite{r82} introduced a multi-tasking framework for image restoration using a degradation-aware VLM. 

\begin{figure*}[t]
  \centering
   \includegraphics[width=1.0\linewidth]{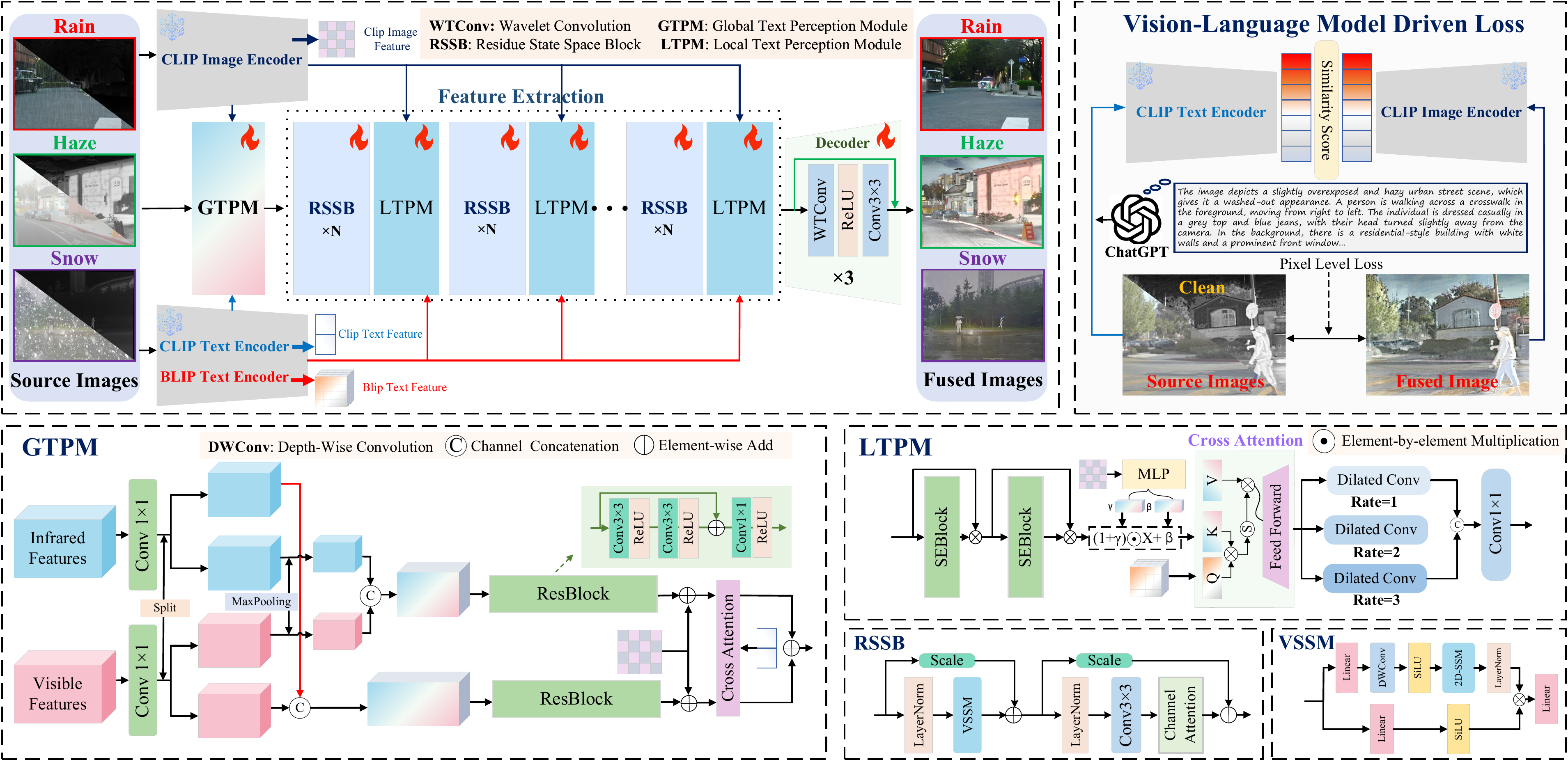}
   \caption{The architecture of the proposed multi-modality image fusion method for adverse weather.}
   \label{fig2}
\end{figure*}

\section{Methodology}
\subsection{Overview.}
This section presents the proposed all-weather multi-modality image fusion method via global and local text perception, with the specific workflow illustrated in Figure~\ref{fig2}.

Our method consists of four main components: (1) \textbf{G}lobal \textbf{T}ext \textbf{P}erception \textbf{M}odule (\textbf{GTPM}), (2) \textbf{R}esidual \textbf{S}tate \textbf{S}pace \textbf{B}lock (\textbf{RSSB}) \cite{r95}, (3) \textbf{L}ocal \textbf{T}ext \textbf{P}erception \textbf{M}odule (\textbf{LTPM}), and (4) the decoder based on wavelet convolution.
The GTPM is designed to establish a deep correlation between image features and global textual features, enhancing the ability of our model  to perceive and adapt to various degraded scenes. Mamba provides a potential solution for balancing the global receptive field with computational efficiency, enabling the modeling of long-range dependencies with minimal computational cost \cite{r96,r97}. We use the RSSB as an intermediate feature extraction module, effectively addressing spatial information loss and channel redundancy issues introduced by the flattening strategy in the traditional Mamba module \cite{r95}.
The LTPM, positioned directly after the RSSB, enhances the contextual understanding of model by guiding it with detailed semantic information. This approach ensures semantic consistency in the fused features while improving the extraction of finer details.

The image decoder must quickly reconstruct the holistic content of the image. However, small kernel convolutions (e.g., $3$$\times$$3$) limit the receptive field of CNN. Inspired by large-kernel convolutional networks \cite{r100,r101,r102}, we find that increasing the kernel size is more effective than stacking additional layers. Thus, we incorporate wavelet convolution (WTConv) \cite{r103} in the decoder. By extracting features in the wavelet domain, WTConv effectively enlarges the receptive field without significantly increasing the parameter count.

\subsection{Global Text Perception Module}
To enhance the ability of proposed algorithm to handle various degraded scenes, we employ the GTPM, which accepts captions generated by BLIP as text input and integrates features encoded from the multi-modality source image via CLIP to supplement scene features. The main process of GTPM is illustrated in \cref{fig2}. First, a $1 $$\times$$ 1$ convolution expands the feature channels of the source image to obtain the visible feature $Img_{vi}$ and the infrared feature $Img_{ir}$. This expansion not only increases the representational capacity of the feature space but also facilitates subsequent feature interaction. Next, the infrared and visible feature maps are split into two parts for multi-scale fusion. 
\begin{equation}
  \text{\textit{Img}}_m^L, \ \text{\textit{Img}}_m^R = \text{Split}(\text{\textit{Img}}_m)
  \label{eq1}
\end{equation}
where $Img_{m}$ represents the image features with $m=vi$ for visible and $m=ir$ for infrared, $L$ and $R$ denote the left and right halves respectively.
For the left half, infrared and visible features are concatenated directly to produce the first set of fused features $\text{Fuse}^L$, while the right half undergoes max pooling before concatenation to obtain the second set $\text{Fuse}^R$ , thereby capturing hierarchical and multi-scale contextual information. This decomposition allows the model to simultaneously leverage fine-grained details and broader contextual dependencies, enhancing the robustness of fusion across diverse degradation patterns. This operation can be formally expressed as follows:
\begin{equation}
  \text{\textit{Fuse}}^L = \text{Concat}(\text{\textit{Img}}_{vi}^L, \text{\textit{Img}}_{ir}^L)
  \label{eq2}
\end{equation}
\begin{equation}
  \text{\textit{Fuse}}^R = \text{MAP}(\text{Concat}(\text{\textit{Img}}_{vi}^R, \text{\textit{Img}}_{ir}^R))
  \label{eq3}
\end{equation} 
where $\text{MAP}(\cdot)$ represents max pooling function. These two sets of fused features are then processed by a residual block to enhance feature representation and minimize feature loss. The residual block effectively mitigates potential gradient vanishing issues during fusion and improves feature robustness. The fused features are subsequently integrated with image features extracted by the CLIP encoder, ensuring consistency across multi-modality information. To incorporate semantic information and improve model comprehension, we employ a cross-attention module \cite{r80} to fuse text and image features. In this setup, text features serve as queries $Q_{\text{\textit{Text}}}$, while image features act as keys $K_{\text{Fuse}}$ and values $V_{\text{Fuse}}$, enabling the model to adaptively locate the relevant image regions corresponding to each text feature. It can be expressed as follows:
\begin{equation}
\mathrm{CA}(F_{\text{\textit{Text}}}, \text{\textit{Fuse}}) = \text{Softmax} \left( \frac{Q_{\text{\textit{Text}}} K_{\text{\textit{Fuse}}}}{\sqrt{d_k}} \right) V_{\text{\textit{Fuse}}}
  \label{eq4}
\end{equation}
where $\mathrm{CA}(\cdot)$ and $\sqrt{d_k}$ are the cross-attention function and the dimension of the key, respectively, $F_{\text{\textit{Text}}}$ denotes the text feature.

\subsection{Local Text Perception Module}
To capture finer details in degraded scenes and more accurately restore original pixel information, we introduce LTPM and integrate it into the feature extraction process of the network. Figure~\ref{fig2} illustrates the detailed workflow of LTPM.
The LTPM module first takes input from the fused features and processes it through a two-layer SEBlock, enhancing local feature representation by emphasizing important features and reducing redundant information via an adaptive channel attention mechanism. Drawing on the Text-IF \cite{r65}, we incorporate the image features generated by the CLIP encoder with the current fused features through scaling and bias control. Specifically, the CLIP-encoded image features are projected into the semantic space of the fused features through an MLP layer, effectively embedding the global image information into the fusion process. Assuming the scale scaling and bias control parameters obtained from the MLP are denoted as $\gamma$ and $\beta$, the process can be represented as follows:
\begin{equation}
\hat{Fuse} = (1 + \gamma) \odot X + \beta
  \label{eq5}
\end{equation}
where $\odot$ represents the element-by-element multiplication operation and $X$ represents the fusion features obtained by RSSB.
The processed features are then input into a cross-attention module, accompanied by detailed textual features encoded via BLIP, to enhance the integration of multi-modality information from both visual and textual sources. This cross-attention mechanism maps image and text data into a unified feature space, effectively capturing their deep relationships. Given the potential redundancy from the added text details, the module employs dilated convolution to extract multi-scale features at three different dilation rates. Finally, these three layers of multi-scale features are concatenated to form the output of the module.

\begin{figure*}[t]
  \centering
   \includegraphics[width=1\linewidth]{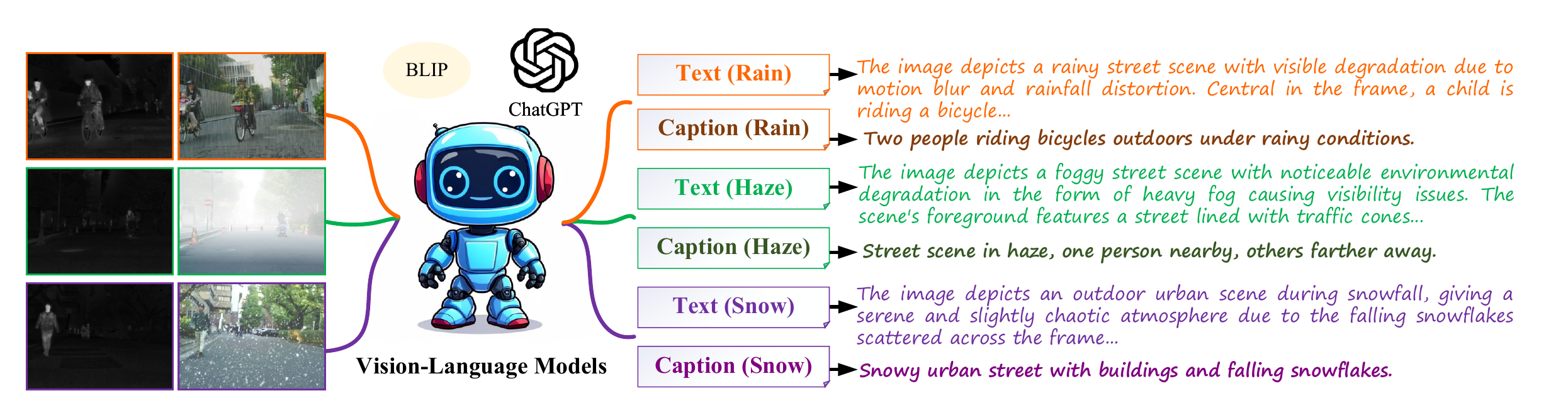}
   \caption{Text descriptions of different multi-modality images in adverse weather conditions.}
   \label{text}
\end{figure*}

\subsection{Loss Functions}
Due to the strong semantic perception capabilities of VLMs, some prior studies \cite{r105,r106} have leveraged them to guide model training. For example, Xu et al. \cite{r105} employed CLIP to introduce a weather-prompt learning loss, while Chen et al. \cite{r106} proposed an image-perception prompt learning strategy. However, these approaches \cite{r105,r106} are limited to image restoration and enhancement tasks, lacking generalization to IVIF, and there is a lack of clean semantic text descriptions corresponding to multi-modality images, particularly for those captured under adverse weather conditions.

To enhance the semantic features of fused images, we propose a novel VLM-driven loss. As shown in Figure~\ref{fig2}, this method inputs the fusion result into the CLIP image encoder, while the clean multi-modality images description is processed by the text encoder, aligning visual and textual features within a shared feature space. By calculating the cosine similarity between these two feature sets, the method effectively measures content consistency between the image and text. The VLM-driven loss $\mathcal{L}_{\text{{VLM}}}$ is expressed as follows:
\begin{equation}
\mathcal{L}_{\text{{VLM}}} = 1 - \frac{F_{\text{\textit{Img}}} \cdot F_{\text{\textit{Text}}}^{\text{\textit{Clean}}}}{\|F_{\text{\textit{Img}}}\| \|F_{\text{\textit{Text}}}^{\text{\textit{Clean}}}\|}
  \label{eq6}
\end{equation}
where $F_{\text{\textit{Img}}}$ and $F_{\textit{Text}}^{\textit{Clean}}$ denote the features obtained from fusion results and text for clean multi-modality images after CLIP encoding respectively, and $ \| \cdot \|$ represents the norm.

In addition, we introduce pixel-level losses to ensure that the network generates high-fidelity fusion results. These include color consistency loss $\mathcal{L}_{\text{Color}}$, L1 loss $\mathcal{L}_{\text{L1}}$, and structural similarity measure loss $\mathcal{L}_{\text{SSIM}}$. Letting ${\textit{Img}}_F$ represent the fusion result and ${\textit{Img}}_{m}^{\text{\textit{Clean}}}$ represent the clean multi-modality images, the loss $\mathcal{L}_{\text{Color}}$ can be expressed as follows:
\begin{equation}
\mathcal{L}_{\text{Color}} = \frac{1}{HW} \left\| \text{T}_{\text{CbCr}}(\mathrm{\textit{Img}}_F) - \text{T}_{\text{CbCr}}(\mathrm{\textit{Img}}_{vi}^{\text{\textit{Clean}}}) \right\|_1
  \label{eq7}
\end{equation}
where $\text{T}_{\text{CbCr}}$ denotes the transfer function of RGB to CbCr. $\mathcal{L}_{\text{L1}}$ can be expressed as follows:
\begin{equation}
\mathcal{L}_{\text{L1}} = \frac{1}{HW} \sum_{m \in \{\text{ir}, vi\}} \left\| \mathrm{\textit{Img}}_F - \mathrm{\textit{Img}}_{m}^{\text{\textit{Clean}}} \right\|_1
  \label{eq8}
\end{equation}
$\mathcal{L}_{\text{SSIM}}$ can be expressed as follows:
\begin{equation}
\mathcal{L}_{\text{SSIM}} = 2 - \sum_{m \in \{v_i, \text{ir}\}} \text{SSIM}(\mathrm{\textit{Img}}_F, \mathrm{\textit{Img}}_{m}^{\text{\textit{Clean}}})
  \label{eq9}
\end{equation}
where $\text{SSIM}(\cdot)$ denotes the structural similarity measurement operator. The total loss $\mathcal{L}_{\text{Total}}$ is calculated as follows:
\begin{equation}
\mathcal{L}_{\text{Total}} = \mathcal{L}_{\text{{VLM}}} + \mathcal{L}_{\text{Color}} + \mathcal{L}_{\text{L1}} + \mathcal{L}_{\text{SSIM}}
  \label{eq10}
\end{equation}

\subsection{Text Quality Control for Multi-Modality Images}
To ensure high-quality textual supervision for our fusion framework, we control the generation and selection of both detailed and global text descriptions. Detailed scene descriptions are generated by ChatGPT-4 and adjusted to comply with the maximum text length that CLIP can encode, while global captions are produced by BLIP to fit its own text encoding constraints. The quality of all generated texts is primarily verified through manual screening. In total, we generated detailed and global textual descriptions for 8,500 multi-modality images captured under adverse weather conditions, along with their corresponding clean images.

For quality assurance, we adopt a stratified random sampling strategy, randomly selecting approximately 30\% of the generated texts for manual inspection. Each sampled text is evaluated across multiple criteria to ensure accuracy and reliability for subsequent fusion learning. These criteria include semantic consistency, completeness of scene description, adherence to length constraints, and modality-specific relevance. Semantic consistency ensures that the text accurately reflects the content of the source images, including key objects, actions, and environmental features. Modality-specific relevance assesses whether the descriptions provide meaningful cues for each modality, emphasizing thermal information for infrared images and color or texture information for visible images.
Figure~\ref{text} illustrates textual descriptions corresponding to multi-modality images captured under severe weather conditions.


\begin{figure*}[t]
  \centering
   \includegraphics[width=1\linewidth]{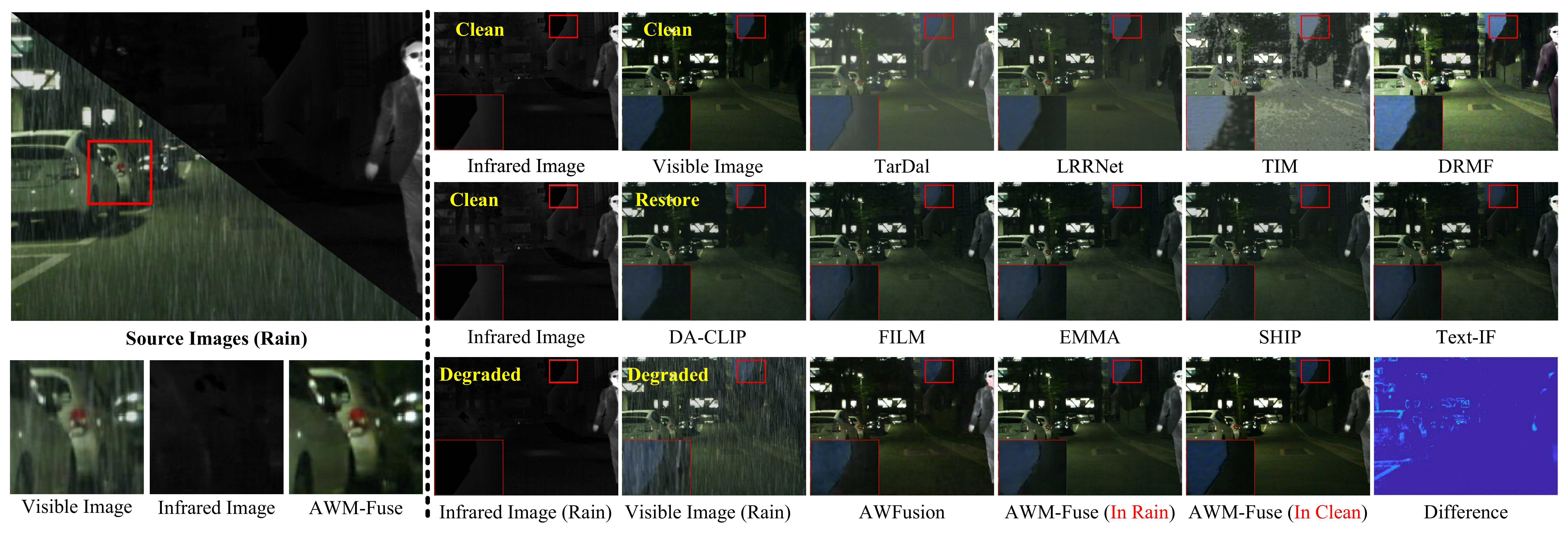}
   \caption{Comparison of image fusion results of different methods in ideal and rain scenes. The "Difference" represents the difference map between AWM-Fuse (In clean) and AWM-Fuse (In Rain).}
   \label{fig3}
\end{figure*}
\begin{figure*}[t]
  \centering
   \includegraphics[width=1\linewidth]{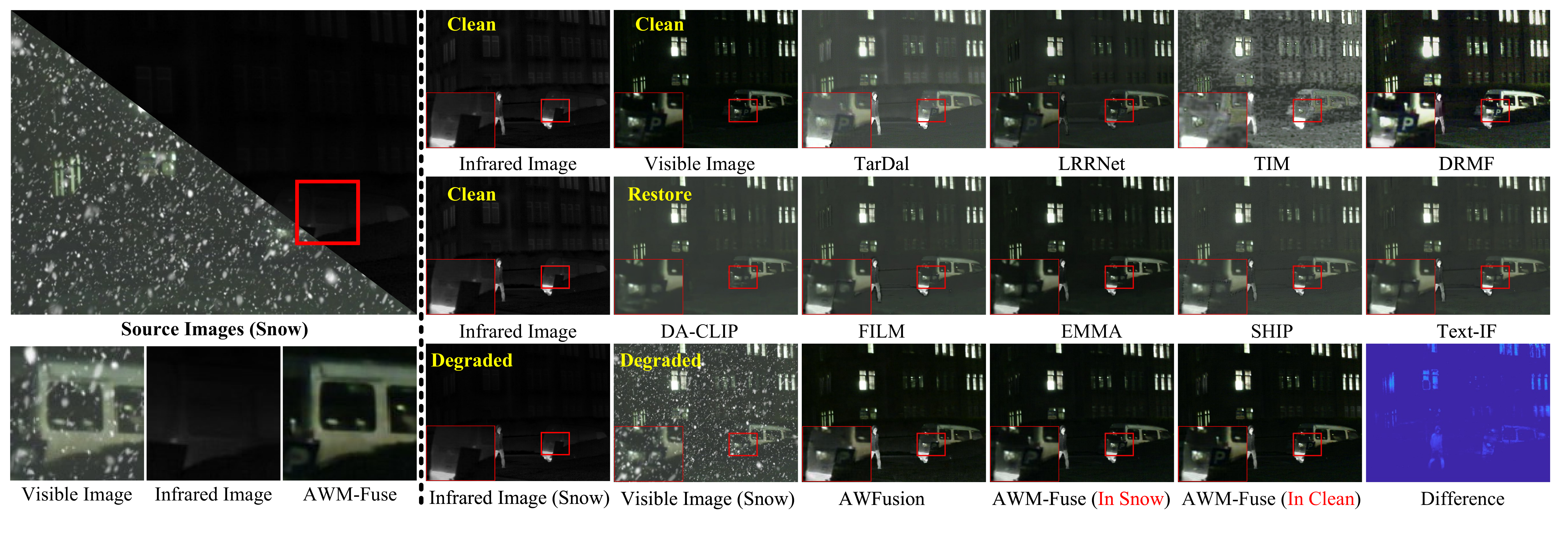}
   \caption{Comparison of image fusion results of different methods in ideal and snow scenes. The "Difference" represents the difference map between AWM-Fuse (In clean) and AWM-Fuse (In Snow).}
   \label{fig4}
\end{figure*}

\section{Experiment}
\subsection{Experimental Setups}
\noindent \textbf{Dataset and Comparison Methods.}
We selected 50 images for each of the three adverse weather types, rain, haze, and snow, from the AWMM-100k dataset\footnote{\url{https://ixilai.github.io/AWMM-100K/}} \cite{r91} to evaluate AWM-Fuse. AWMM-100k contains both real and synthetic samples, providing a comprehensive benchmark for evaluating the  ability of different methods to cope with severe weather.
We compared the fusion results qualitatively and quantitatively with nine state-of-the-art methods, including TarDal \cite{r17}, LRRNet \cite{r32}, TIM \cite{r79}, DRMF \cite{r89}, FILM \cite{r80}, EMMA \cite{r77}, SHIP \cite{r88}, Text-IF \cite{r65} and AWFusion \cite{r91}. TarDal, LRRNet, TIM, and DRMF use clean visible and infrared images as input, while FILM, EMMA, SHIP, and Text-IF rely on visible images restored by two baselines (DehazeFormer \cite{r98} and DA-CLIP \cite{r82}) and clean infrared images. Although DRMF and Text-IF are designed to complex scenarios, they are not applicable to adverse weather, so we evaluated them  in a comparative manner consistent with other normal fusion methods. In contrast, AWFusion and AWM-Fuse processes degraded visible and infrared images directly.

In the ideal scenes, we conducted experiments on three standard datasets ( M3FD \cite{r17}, MSRS \cite{r62} and LLVIP \cite{r76}). Since these datasets do not contain degradation, all methods are directly based on the original multi-modal images for fusion comparisons.


\noindent \textbf{Metrics.}
The metrics we use include the Gradient-Based Metric $Q_{G}$, the Image Fusion Metric-Based metric $Q_{M}$, Piella’s Metric $Q_{S}$, the Chen-Varshney Metric $Q_{CV}$ \cite{r69}, the Sum of the Correlation of Differences $SCD$ \cite{r70}, Visual Information Fidelity $VIF$ \cite{r99}, and the Structural Similarity Index Measure $SSIM$. Except for $Q_{CV}$, higher scores on these metrics indicate better quality.

\noindent \textbf{Training Setups.}
During training, we extracted 700 images for each type of adverse weather (rain, haze, and snow) from the AWMM-100k dataset, combining them into a single dataset of 2100 image pairs. Each image was randomly cropped to $160 $$\times$$ 160$ pixels, with a batch size of 2. The Adam optimizer was used with an initial learning rate of $1$$\times$$10^{-3}$, and the model was trained for over 300 epochs. For network configuration, our RSSBs employ a U-Net structure, with the block configuration of each layer set to $[8, 10, 10, 12, 10, 10, 8]$, achieving a balance between performance and parameter efficiency. The number of feature channels in the RSSBs progressively increase from 48 to 384 and is then reduced back to 48 channels.

\begin{figure*}[t]
  \centering
    \includegraphics[width=1\linewidth]{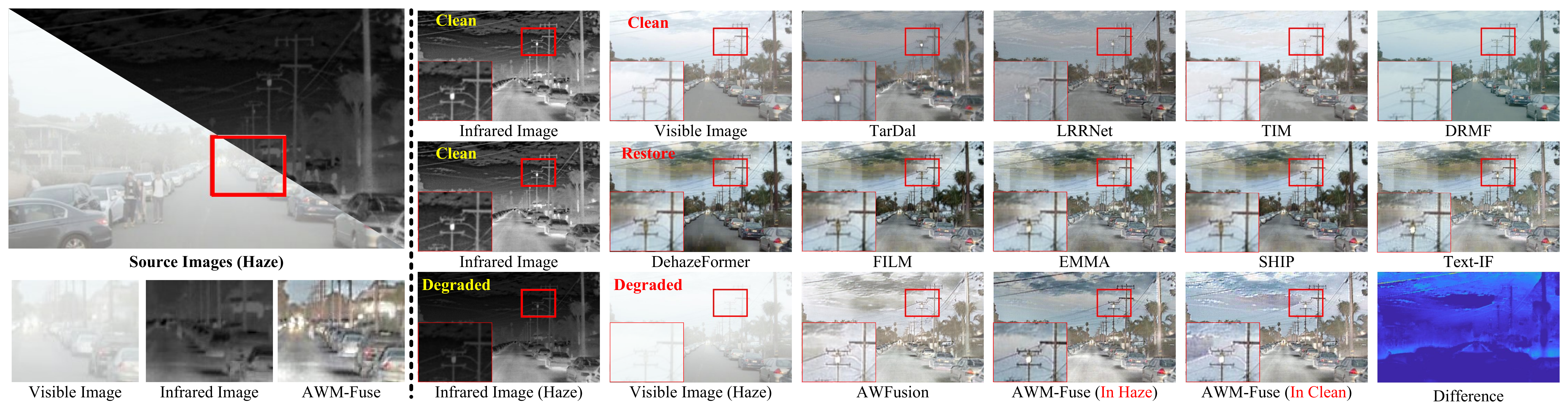}
  \caption{Comparison of image fusion results of different methods in ideal and haze scenes. The "Difference" represents the difference map between AWM-Fuse (In clean) and AWM-Fuse (In Haze).}
  \label{fig5}
\end{figure*}
\begin{table*}[t]
\caption{Comparison of quantitative results of different methods in ideal and adverse weather scenes. The best scores are in bold, while the second-best scores are in blue.}
\begin{adjustbox}{width=\textwidth}
\centering
\tiny
\begin{tabular}{ccccccccccc}
\hline
\multicolumn{1}{c|}{Source}                 & \multicolumn{1}{c|}{Methods}    & \multicolumn{1}{c|}{Pub.}                & \multicolumn{1}{c|}{Baseline}     & $Q_G$ $\uparrow$                                    & $Q_M$ $\uparrow$                                      & $Q_S$ $\uparrow$                                      & $Q_{CV}$ $\downarrow$                                       & \textit{VIF}$\uparrow$                                      & \textit{SSIM}$\uparrow$                                     & \textit{SCD}$\uparrow$                                      \\ \hline
\multicolumn{1}{c|}{}                       & \multicolumn{1}{c|}{TarDal \cite{r17}}     & \multicolumn{1}{c|}{\textit{CVPR 2022}}  & \multicolumn{1}{c|}{$\times$}            & 0.3563                                 & 0.3859                                 & 0.5739                                 & 653.6011                                 & 0.3010                                 & 0.3053                                 & 1.5556                                 \\
\multicolumn{1}{c|}{}                       & \multicolumn{1}{c|}{LRRNet \cite{r32}}     & \multicolumn{1}{c|}{\textit{TPAMI 2023}} & \multicolumn{1}{c|}{$\times$}            & 0.3194                                 & 0.5046                                 & 0.7003                                 & 682.5369                                 & {\color[HTML]{0000FF} {0.3388}} & 0.1708                                 & 0.7897                                 \\
\multicolumn{1}{c|}{}                       & \multicolumn{1}{c|}{TIM \cite{r79}}        & \multicolumn{1}{c|}{\textit{TPAMI 2024}} & \multicolumn{1}{c|}{$\times$}            & 0.2057                                 & 0.3498                                 & 0.4630                                 & 827.1437                                 & 0.1221                                 & 0.1585                                 & 0.9651                                 \\
\multicolumn{1}{c|}{\multirow{-4}{*}{Clean}}   & \multicolumn{1}{c|}{DRMF \cite{r89}}       & \multicolumn{1}{c|}{\textit{MM 2024}}    & \multicolumn{1}{c|}{$\times$}            & 0.3108                                 & 0.4170                                 & 0.5844                                 & 929.2306                                 & 0.2579                                 & 0.2743                                 & 0.8937                                 \\ \hline
\multicolumn{1}{c|}{}                       & \multicolumn{1}{c|}{FILM \cite{r80}}       & \multicolumn{1}{c|}{\textit{ICML 2024}}  & \multicolumn{1}{c|}{DA-CLIP }      & 0.3522                                 & 0.5097                                 & 0.7963                                 & 560.7920                                 & 0.2512                                 & 0.3047                                 & 1.3869                                 \\
\multicolumn{1}{c|}{}                       & \multicolumn{1}{c|}{EMMA \cite{r77}}       & \multicolumn{1}{c|}{\textit{CVPR 2024}}  & \multicolumn{1}{c|}{DA-CLIP }      & 0.2812                                 & 0.4131                                 & 0.8019                                 & 493.5733                                 & 0.2418                                 & 0.3013                                 & 1.4011                                 \\
\multicolumn{1}{c|}{}                       & \multicolumn{1}{c|}{SHIP \cite{r88}}       & \multicolumn{1}{c|}{\textit{CVPR 2024}}  & \multicolumn{1}{c|}{DA-CLIP }      & 0.3139                                 & 0.4105                                 & 0.7575                                 & 584.6434                                 & 0.2354                                 & 0.2805                                 & 1.2948                                 \\
\multicolumn{1}{c|}{}                       & \multicolumn{1}{c|}{Text-IF \cite{r65}}    & \multicolumn{1}{c|}{\textit{CVPR 2024}}  & \multicolumn{1}{c|}{DA-CLIP }      & {\color[HTML]{0000FF} {0.3771}} & 0.4645                                 & 0.8231                                 & 549.9020                                 & 0.2784                                 & {\color[HTML]{0000FF} {0.3591}} & 1.4088                                 \\ \cline{2-11} 
\multicolumn{1}{c|}{}                       & \multicolumn{1}{c|}{AWFusion \cite{r91}}   & \multicolumn{1}{c|}{\textit{Arxiv 2024}}      & \multicolumn{1}{c|}{$\times$}            & 0.3334                                 & {\color[HTML]{0000FF} {0.6428}} & {\color[HTML]{0000FF} {0.8335}} & {\color[HTML]{0000FF} {476.3436}} & 0.2997                                 & 0.3528                                 & {\color[HTML]{0000FF} {1.6010}} \\
\multicolumn{1}{c|}{\multirow{-6}{*}{Rain}} & \multicolumn{1}{c|}{AWM-Fuse} & \multicolumn{1}{c|}{-}                   & \multicolumn{1}{c|}{$\times$}            & \textbf{0.4321}                        & \textbf{0.7209}                        & \textbf{0.8538}                        & \textbf{452.2425}                        & \textbf{0.3574}                        & \textbf{0.4156}                        & \textbf{1.6527}                        \\ \hline
                                            &                                 &                                          &                                   &                                        &                                        &                                        &                                          &                                        &                                        &                                        \\ \hline
\multicolumn{1}{c|}{}                       & \multicolumn{1}{c|}{TarDal \cite{r17}}     & \multicolumn{1}{c|}{\textit{CVPR 2022}}  & \multicolumn{1}{c|}{$\times$}            & 0.4109                                 & 0.3832                                 & 0.7308                                 & {\color[HTML]{0000FF} {337.3376}} & 0.3336                                 & 0.4009                                 & \textbf{1.4207}                        \\
\multicolumn{1}{c|}{}                       & \multicolumn{1}{c|}{LRRNet \cite{r32}}     & \multicolumn{1}{c|}{\textit{TPAMI 2023}} & \multicolumn{1}{c|}{$\times$}            & 0.4154                                 & 0.4775                                 & 0.7646                                 & 461.4354                                 & {\color[HTML]{0000FF} {0.3736}} & 0.3046                                 & 0.8130                                 \\
\multicolumn{1}{c|}{}                       & \multicolumn{1}{c|}{TIM \cite{r79}}        & \multicolumn{1}{c|}{\textit{TPAMI 2024}} & \multicolumn{1}{c|}{$\times$}            & 0.3020                                 & 0.3983                                 & 0.6499                                 & 507.1376                                 & 0.1834                                 & 0.2833                                 & 0.8978                                 \\
\multicolumn{1}{c|}{\multirow{-4}{*}{Clean}}   & \multicolumn{1}{c|}{DRMF \cite{r89}}       & \multicolumn{1}{c|}{\textit{MM 2024}}    & \multicolumn{1}{c|}{$\times$}            & {\color[HTML]{0000FF} {0.5203}} & {\color[HTML]{0000FF} {0.6138}} & 0.7530                                 & 367.8078                                 & 0.3586                                 & {\color[HTML]{0000FF} {0.4182}} & 1.0414                                 \\ \hline
\multicolumn{1}{c|}{}                       & \multicolumn{1}{c|}{FILM \cite{r80}}       & \multicolumn{1}{c|}{\textit{ICML 2024}}  & \multicolumn{1}{c|}{DehazeFormer } & 0.4452                                 & 0.4892                                 & 0.7923                                 & 395.0446                                 & 0.2803                                 & 0.3880                                 & 1.1668                                 \\
\multicolumn{1}{c|}{}                       & \multicolumn{1}{c|}{EMMA \cite{r77}}       & \multicolumn{1}{c|}{\textit{CVPR 2024}}  & \multicolumn{1}{c|}{DehazeFormer } & 0.3993                                 & 0.4341                                 & 0.7782                                 & 386.5206                                 & 0.2749                                 & 0.3869                                 & 1.1894                                 \\
\multicolumn{1}{c|}{}                       & \multicolumn{1}{c|}{SHIP \cite{r88}}       & \multicolumn{1}{c|}{\textit{CVPR 2024}}  & \multicolumn{1}{c|}{DehazeFormer } & 0.4200                                 & 0.4347                                 & 0.7753                                 & 388.9155                                 & 0.2697                                 & 0.3736                                 & 1.1580                                 \\
\multicolumn{1}{c|}{}                       & \multicolumn{1}{c|}{Text-IF \cite{r65}}    & \multicolumn{1}{c|}{\textit{CVPR 2024}}  & \multicolumn{1}{c|}{DehazeFormer } & 0.4422                                 & 0.4706                                 & {\color[HTML]{0000FF} {0.7979}} & 381.8256                                 & 0.2895                                 & 0.4062                                 & 1.2206                                 \\ \cline{2-11} 
\multicolumn{1}{c|}{}                       & \multicolumn{1}{c|}{AWFusion \cite{r91}}   & \multicolumn{1}{c|}{\textit{Arxiv 2024}}      & \multicolumn{1}{c|}{$\times$}            & 0.4283                                 & 0.5471                                 & 0.7738                                 & 462.8585                                 & 0.3433                                 & 0.3901                                 & 1.1719                                 \\
\multicolumn{1}{c|}{\multirow{-6}{*}{Haze}} & \multicolumn{1}{c|}{AWM-Fuse} & \multicolumn{1}{c|}{-}                   & \multicolumn{1}{c|}{$\times$}            & \textbf{0.5297}                        & \textbf{0.8193}                        & \textbf{0.8392}                        & \textbf{168.0220}                        & \textbf{0.3936}                        & \textbf{0.4631}                        & {\color[HTML]{0000FF} {1.3416}} \\ \hline
\multicolumn{1}{l}{}                        & \multicolumn{1}{l}{}            & \multicolumn{1}{l}{}                     & \multicolumn{1}{l}{}              & \multicolumn{1}{l}{}                   & \multicolumn{1}{l}{}                   & \multicolumn{1}{l}{}                   & \multicolumn{1}{l}{}                     & \multicolumn{1}{l}{}                   & \multicolumn{1}{l}{}                   & \multicolumn{1}{l}{}                   \\ \hline
\multicolumn{1}{c|}{}                       & \multicolumn{1}{c|}{TarDal \cite{r17}}     & \multicolumn{1}{c|}{\textit{CVPR 2022}}  & \multicolumn{1}{c|}{$\times$}            & 0.3811                                 & 0.3718                                 & 0.6238                                 & 553.2230                                 & {\color[HTML]{0000FF} {0.3142}} & 0.3453                                 & 1.4684                                 \\
\multicolumn{1}{c|}{}                       & \multicolumn{1}{c|}{LRRNet \cite{r32}}     & \multicolumn{1}{c|}{\textit{TPAMI 2023}} & \multicolumn{1}{c|}{$\times$}            & 0.3574                                 & 0.4651                                 & 0.7154                                 & 661.1762                                 & 0.3420                                 & 0.2361                                 & 0.8301                                 \\
\multicolumn{1}{c|}{}                       & \multicolumn{1}{c|}{TIM \cite{r79}}        & \multicolumn{1}{c|}{\textit{TPAMI 2024}} & \multicolumn{1}{c|}{$\times$}            & 0.2433                                 & 0.3626                                 & 0.5181                                 & 820.5891                                 & 0.1505                                 & 0.2105                                 & 0.8977                                 \\
\multicolumn{1}{c|}{\multirow{-4}{*}{Clean}}   & \multicolumn{1}{c|}{DRMF \cite{r89}}       & \multicolumn{1}{c|}{\textit{MM 2024}}    & \multicolumn{1}{c|}{$\times$}            & {\color[HTML]{0000FF} {0.3948}} & 0.4684                                 & 0.6747                                 & 729.0171                                 & 0.2827                                 & 0.3417                                 & 1.2065                                 \\ \hline
\multicolumn{1}{c|}{}                       & \multicolumn{1}{c|}{FILM \cite{r80}}       & \multicolumn{1}{c|}{\textit{ICML 2024}}  & \multicolumn{1}{c|}{DA-CLIP }      & 0.2959                                 & 0.4299                                 & 0.6296                                 & 504.9559                                 & 0.2189                                 & 0.2491                                 & 1.2575                                 \\
\multicolumn{1}{c|}{}                       & \multicolumn{1}{c|}{EMMA \cite{r77}}       & \multicolumn{1}{c|}{\textit{CVPR 2024}}  & \multicolumn{1}{c|}{DA-CLIP }      & 0.2490                                 & 0.3644                                 & 0.7310                                 & 462.8019                                 & 0.1934                                 & 0.2472                                 & 1.3021                                 \\
\multicolumn{1}{c|}{}                       & \multicolumn{1}{c|}{SHIP \cite{r88}}       & \multicolumn{1}{c|}{\textit{CVPR 2024}}  & \multicolumn{1}{c|}{DA-CLIP }      & 0.3400                                 & 0.3601                                 & 0.5625                                 & 505.2532                                 & 0.2158                                 & 0.2417                                 & 1.1179                                 \\
\multicolumn{1}{c|}{}                       & \multicolumn{1}{c|}{Text-IF \cite{r65}}    & \multicolumn{1}{c|}{\textit{CVPR 2024}}  & \multicolumn{1}{c|}{DA-CLIP }      & 0.3508                                 & 0.4162                                 & 0.6444                                 & 448.5905                                 & 0.2549                                 & 0.2906                                 & 1.3108                                 \\ \cline{2-11} 
\multicolumn{1}{c|}{}                       & \multicolumn{1}{c|}{AWFusion \cite{r91}}   & \multicolumn{1}{c|}{\textit{Arxiv 2024}}      & \multicolumn{1}{c|}{$\times$}            & 0.3552                                 & {\color[HTML]{0000FF} {0.5535}} & {\color[HTML]{0000FF} {0.8119}} & {\color[HTML]{0000FF} {270.1595}} & 0.3024                                 & {\color[HTML]{0000FF} {0.3734}} & {\color[HTML]{0000FF} {1.4937}} \\
\multicolumn{1}{c|}{\multirow{-6}{*}{Snow}} & \multicolumn{1}{c|}{AWM-Fuse} & \multicolumn{1}{c|}{-}                   & \multicolumn{1}{c|}{$\times$}            & \textbf{0.4499}                        & \textbf{0.7046}                        & \textbf{0.8456}                        & \textbf{217.7734}                        & \textbf{0.3573}                        & \textbf{0.4406}                        & \textbf{1.6263}                        \\ \hline
\end{tabular}
\end{adjustbox}
\label{tab1}
\end{table*}

\begin{table*}[t]
\caption{Quantitative comparison  of all methods in three standard datasets. The best scores are in bold, while the second-best scores are in blue.}
\centering
\footnotesize
\begin{tabular}{cccccccccc}
\hline
\multicolumn{1}{c|}{Datasets}               & \multicolumn{1}{c|}{Methods}  & \multicolumn{1}{c|}{Pub.}       & $Q_{G}$$\uparrow$                   & $Q_{M}$$\uparrow$                  & $Q_{S}$$\uparrow$                   & $Q_{CV}$$\downarrow$                  & $VIF$$\uparrow$                  & $SSIM$$\uparrow$               & $SCD$$\uparrow$               \\ \hline
\multicolumn{1}{c|}{\multirow{9}{*}{M3FD}}  & \multicolumn{1}{c|}{TarDal \cite{r17}}   & \multicolumn{1}{c|}{\textit{CVPR 2022}}  & 0.3713               & 0.3903               & 0.8214               & 566.8158             & 0.3205               & 0.4568               & \textbf{1.5203}      \\
\multicolumn{1}{c|}{}                       & \multicolumn{1}{c|}{LRRNet \cite{r32}}   & \multicolumn{1}{c|}{\textit{TPAMI 2023}} & 0.3641               & 0.4823               & 0.8395               & 759.9919             & 0.4027               & 0.3833               & 1.3502               \\
\multicolumn{1}{c|}{}                       & \multicolumn{1}{c|}{TIM \cite{r79}}      & \multicolumn{1}{c|}{\textit{TPAMI 2024}} & 0.3447               & 0.4325               & 0.7928               & 1135.7748            & 0.2678               & 0.3553               & 0.4246               \\
\multicolumn{1}{c|}{}                       & \multicolumn{1}{c|}{DRMF  \cite{r89}}     & \multicolumn{1}{c|}{\textit{MM 2024}}    & 0.3420               & 0.4707               & 0.8041               & 975.7381             & 0.3808               & 0.3493               & 0.8389               \\
\multicolumn{1}{c|}{}                       & \multicolumn{1}{c|}{FILM \cite{r80}}     & \multicolumn{1}{c|}{\textit{ICML 2024}}  & 0.4892               & 1.0778               & 0.8559               & 526.8890             & 0.3840               & 0.4568               & 1.2927               \\
\multicolumn{1}{c|}{}                       & \multicolumn{1}{c|}{EMMA \cite{r77}}     & \multicolumn{1}{c|}{\textit{CVPR 2024}}  & 0.4665               & 0.5598               & 0.8501               & 529.0906             & 0.3839               & 0.4644               & 1.3918               \\
\multicolumn{1}{c|}{}                       & \multicolumn{1}{c|}{SHIP  \cite{r88}}     & \multicolumn{1}{c|}{\textit{CVPR 2024}}  & \color[HTML]{0000FF}{0.5572}               &\color[HTML]{0000FF}{1.5184 }                   & 0.8492               & 519.7535             & 0.3897               & 0.4476               & 1.1945               \\
\multicolumn{1}{c|}{}                       & \multicolumn{1}{c|}{Text-IF \cite{r65}}  & \multicolumn{1}{c|}{\textit{CVPR 2024}}  & 0.5344               & 1.0641               & \textbf{0.8724}      & \textbf{448.7277}    & \color[HTML]{0000FF}{0.4173}               &\color[HTML]{0000FF}{0.4901}               & \color[HTML]{0000FF}{1.4025}               \\
\multicolumn{1}{c|}{}                       & \multicolumn{1}{c|}{AWFusion \cite{r91}}  & \multicolumn{1}{c|}{\textit{Arxiv 2024}}  & 0.4538
              & 0.5680
               & 0.8420
      & 544.1436
    & 0.3723
              &0.4657
               & 1.1963
              \\
\multicolumn{1}{c|}{}                       & \multicolumn{1}{c|}{\cellcolor[HTML]{EFEFEF}AWM-Fuse} & \multicolumn{1}{c|}{\cellcolor[HTML]{EFEFEF}-}          & \textbf{\cellcolor[HTML]{EFEFEF}0.5752}      & \textbf{\cellcolor[HTML]{EFEFEF}1.5261}      & \color[HTML]{0000FF}{\cellcolor[HTML]{EFEFEF}0.8619  }             & \color[HTML]{0000FF}{\cellcolor[HTML]{EFEFEF}503.9105}             & \textbf{\cellcolor[HTML]{EFEFEF}0.4439}      & \textbf{\cellcolor[HTML]{EFEFEF}0.5036}      & \cellcolor[HTML]{EFEFEF}1.2467               \\ \hline
\multicolumn{1}{l}{}                        & \multicolumn{1}{l}{}          & \multicolumn{1}{l}{}            & \multicolumn{1}{l}{} & \multicolumn{1}{l}{} & \multicolumn{1}{l}{} & \multicolumn{1}{l}{} & \multicolumn{1}{l}{} & \multicolumn{1}{l}{} & \multicolumn{1}{l}{} \\ \hline
\multicolumn{1}{c|}{\multirow{9}{*}{MSRS}}  & \multicolumn{1}{c|}{TarDal \cite{r17}}   & \multicolumn{1}{c|}{\textit{CVPR 2022}}  & 0.4109               & 0.3832               & 0.7308               & 337.3376             & 0.3336               & 0.4009               & 1.4207               \\
\multicolumn{1}{c|}{}                       & \multicolumn{1}{c|}{LRRNet \cite{r32}}   & \multicolumn{1}{c|}{\textit{TPAMI 2023}} & 0.4154               & 0.4775               & 0.7646               & 461.4354             & 0.3736               & 0.3046               & 0.8130               \\
\multicolumn{1}{c|}{}                       & \multicolumn{1}{c|}{TIM \cite{r79}}      & \multicolumn{1}{c|}{\textit{TPAMI 2024}} & 0.3020               & 0.3983               & 0.6499               & 507.1376             & 0.1834               & 0.2833               & 0.8978               \\
\multicolumn{1}{c|}{}                       & \multicolumn{1}{c|}{DRMF \cite{r89}}     & \multicolumn{1}{c|}{\textit{MM 2024}}    & 0.5203               & 0.6138               & 0.7530               & 367.8687             & 0.3586               & 0.4182               & 1.0414               \\
\multicolumn{1}{c|}{}                       & \multicolumn{1}{c|}{FILM \cite{r80}}     & \multicolumn{1}{c|}{\textit{ICML 2024}}  & \textbf{0.6715}      & 1.6424               & \color[HTML]{0000FF}{0.8521}               & 148.8272             & 0.4387               & 0.4791               & 1.4297               \\
\multicolumn{1}{c|}{}                       & \multicolumn{1}{c|}{EMMA \cite{r77}}     & \multicolumn{1}{c|}{\textit{CVPR 2024}}  & 0.5851               & 0.7199               & 0.8385               & 140.2975             & 0.4244               & 0.4758               & 1.5062               \\
\multicolumn{1}{c|}{}                       & \multicolumn{1}{c|}{SHIP \cite{r88}}     & \multicolumn{1}{c|}{\textit{CVPR 2024}}  & 0.6280               & \color[HTML]{0000FF}{1.7853}               & 0.8322               & 132.7968             & 0.4166               & 0.4608               & 1.3840               \\
\multicolumn{1}{c|}{}                       & \multicolumn{1}{c|}{Text-IF \cite{r65}}  & \multicolumn{1}{c|}{\textit{CVPR 2024}}  & 0.6465               & 1.4406               & \textbf{0.8546}      & \color[HTML]{0000FF}{130.9396}             & \color[HTML]{0000FF}{0.4487}               & \color[HTML]{0000FF}{0.4909}               & \textbf{1.5675}      \\
\multicolumn{1}{c|}{}                       & \multicolumn{1}{c|}{AWFusion \cite{r91}}  & \multicolumn{1}{c|}{\textit{Arxiv 2024}}  &0.5540
              & 0.8039
              & 0.8466
      & 178.0195
            & 0.4187
               & 0.4775
              & 1.2941
   \\
\multicolumn{1}{c|}{}                       & \multicolumn{1}{c|}{\cellcolor[HTML]{EFEFEF}AWM-Fuse} & \multicolumn{1}{c|}{\cellcolor[HTML]{EFEFEF}-}          & \color[HTML]{0000FF}{\cellcolor[HTML]{EFEFEF}0.6531}               & \textbf{\cellcolor[HTML]{EFEFEF}1.9846}      & \cellcolor[HTML]{EFEFEF}0.8512               & \textbf{\cellcolor[HTML]{EFEFEF}127.9056}    & \textbf{\cellcolor[HTML]{EFEFEF}0.4641}      & \textbf{\cellcolor[HTML]{EFEFEF}0.5042}      & \color[HTML]{0000FF}{\cellcolor[HTML]{EFEFEF}1.5365}               \\ \hline
\multicolumn{1}{l}{}                        & \multicolumn{1}{l}{}          & \multicolumn{1}{l}{}            & \multicolumn{1}{l}{} & \multicolumn{1}{l}{} & \multicolumn{1}{l}{} & \multicolumn{1}{l}{} & \multicolumn{1}{l}{} & \multicolumn{1}{l}{} & \multicolumn{1}{l}{} \\ \hline
\multicolumn{1}{c|}{\multirow{9}{*}{LLVIP}} & \multicolumn{1}{c|}{TarDal \cite{r17}}   & \multicolumn{1}{c|}{\textit{CVPR 2022}}  & 0.3508               & 0.1517               & 0.6742               & 658.5868             & 0.2802               & 0.3761               & 1.4384               \\
\multicolumn{1}{c|}{}                       & \multicolumn{1}{c|}{LRRNet \cite{r32}}   & \multicolumn{1}{c|}{\textit{TPAMI 2023}} & 0.3622               & 0.1866               & 0.7338               & 852.3098             & 0.3569               & 0.3568               & 0.6557               \\
\multicolumn{1}{c|}{}                       & \multicolumn{1}{c|}{TIM \cite{r79}}      & \multicolumn{1}{c|}{\textit{TPAMI 2024}} & 0.3149               & 0.1637               & 0.6415               & 764.2835             & 0.2337               & 0.2906               & 0.7223               \\
\multicolumn{1}{c|}{}                       & \multicolumn{1}{c|}{DRMF \cite{r89}}     & \multicolumn{1}{c|}{\textit{MM 2024}}    & 0.3579               & 0.1576               & 0.6095               & 842.8371             & 0.2363               & 0.3166               & 1.1763               \\
\multicolumn{1}{c|}{}                       & \multicolumn{1}{c|}{FILM \cite{r80}}     & \multicolumn{1}{c|}{\textit{ICML 2024}}  & \color[HTML]{0000FF}{0.6796}               & 1.0252               & \color[HTML]{0000FF}{0.8860}               & 395.3111             & 0.3905               & 0.4842               & 1.5251               \\
\multicolumn{1}{c|}{}                       & \multicolumn{1}{c|}{EMMA \cite{r77}}     & \multicolumn{1}{c|}{\textit{CVPR 2024}}  & 0.5630               & 0.3088               & 0.8639               & 406.5657             & 0.3452               & 0.4822               & \color[HTML]{0000FF}{1.5836}               \\
\multicolumn{1}{c|}{}                       & \multicolumn{1}{c|}{SHIP \cite{r88}}     & \multicolumn{1}{c|}{\textit{CVPR 2024}}  & 0.6628               & \textbf{1.3961}      & 0.8741               & 388.4069             & 0.3792               & 0.4786               & 1.5028               \\
\multicolumn{1}{c|}{}                       & \multicolumn{1}{c|}{Text-IF \cite{r65}}  & \multicolumn{1}{c|}{\textit{CVPR 2024}}  & 0.64790              & 0.8679               & 0.8752               & \textbf{377.1400}    & \textbf{0.4046}      & \color[HTML]{0000FF}{0.4942 }              & 1.5458               \\
\multicolumn{1}{c|}{}                       & \multicolumn{1}{c|}{AWFusion \cite{r91}}  & \multicolumn{1}{c|}{\textit{Arxiv 2024}}  & 0.5843
              & 0.4942
               &0.8701
             & 426.4772
    & 0.3658
      & 0.4854
              & 1.5968
             \\
\multicolumn{1}{c|}{}                       & \multicolumn{1}{c|}{\cellcolor[HTML]{EFEFEF}AWM-Fuse} & \multicolumn{1}{c|}{\cellcolor[HTML]{EFEFEF}-}          & \textbf{\cellcolor[HTML]{EFEFEF}0.6806}      & \color[HTML]{0000FF}{\cellcolor[HTML]{EFEFEF}1.1895}               & \textbf{\cellcolor[HTML]{EFEFEF}0.8937}      & \color[HTML]{0000FF}{\cellcolor[HTML]{EFEFEF}382.5485}             &\color[HTML]{0000FF}{ \cellcolor[HTML]{EFEFEF}0.4014 }              & \textbf{\cellcolor[HTML]{EFEFEF}0.5135}      & \textbf{\cellcolor[HTML]{EFEFEF}1.5889}      \\ \hline
\end{tabular}
\label{tab6}
\end{table*}

\begin{table*}[t]
\caption{Comparing quantitative results of different methods in semantic segmentation task. The best scores are in bold, while the second-best scores are in blue.}
\begin{adjustbox}{width=\textwidth}
\centering
\footnotesize
\begin{tabular}{c|c|c|ccccccccc}
\hline
Source               & Methods  & Baseline     & Background & Car   & Person & Bike  & Curve & Guardrail & Color cone & Bump  & mIoU  \\ \hline
                     & TarDal \cite{r17}   & {$\times$}            & 97.71      & 85.26 & 67.31  & 62.51 & 49.45 & 70.95     & 24.88      & 32.9  & 61.37 \\
                     & LRRNet \cite{r32}   & {$\times$}            & 97.66      & 84.82 & 60.49  & 61.48 & 44.31 & 75.25     & 36.54      & 50.25 & 63.85 \\
                     & TIM \cite{r79}      & {$\times$}            & 97.68      & 82.22 & 65.85  & 58.6  & 38.75 & 75.76     & 53.92      & 51.65 & 65.55 \\
\multirow{-4}{*}{Clean} & DRMF \cite{r89}     & {$\times$}            & 98.31      & 88.11 & 70.5   & 70.54 & 51.75 & 82.46     & 63.97      & 74.93 & 75.07 \\ \hline
                     & FILM \cite{r80} & Dehazeformer  & 97.83      & 85.7  & 69.11  & 65.13 & 52.33 & 76.11     & 32.81      & 40.17 & 64.89 \\
                     & EMMA \cite{r77}     & Dehazeformer  & 97.76      & 85.35 & 66.27  & 64.98 & 53.31 & 75.04     & 31.98      & 37.22 & 63.99 \\
                     & SHIP \cite{r88}     & Dehazeformer  & 98.42      & 88.57 & 72.18  & 69.91 & 63.14 & 79.83     & 62.79      & 72.6  & 75.93 \\
 &
  Text-IF \cite{r65} &
  DehazeFormer  &
  \textbf{98.51} &
  \textbf{89.42} &
  \textbf{73.52} &
  {\color[HTML]{0000FF} 70.69} &
  {\color[HTML]{0000FF} 63.52} &
  79.62 &
  63.76 &
  {\color[HTML]{0000FF} 75.91} &
  {\color[HTML]{0000FF} 76.87} \\ \cline{2-12} 
 &
  AWFusion \cite{r91} &
  {$\times$} &
  98.4 &
  89.03 &
  70.91 &
  69.62 &
  60.08 &
  {\color[HTML]{0000FF} 83.15} &
  {\color[HTML]{0000FF} 64.45} &
  73.72 &
  76.17 \\
\multirow{-6}{*}{Haze} &
  AWM-Fuse &
  {$\times$} &
  {\color[HTML]{0000FF} 98.48} &
  {\color[HTML]{0000FF} 89.09} &
  {\color[HTML]{0000FF} 72.41} &
  \textbf{70.89} &
  \textbf{63.53} &
  \textbf{86.38} &
  \textbf{64.6} &
  \textbf{75.92} &
  \textbf{77.66} \\ \hline
\end{tabular}
\end{adjustbox}
\label{tab2}
\end{table*}

\begin{figure}[t]
  \centering
   \includegraphics[width=1\linewidth]{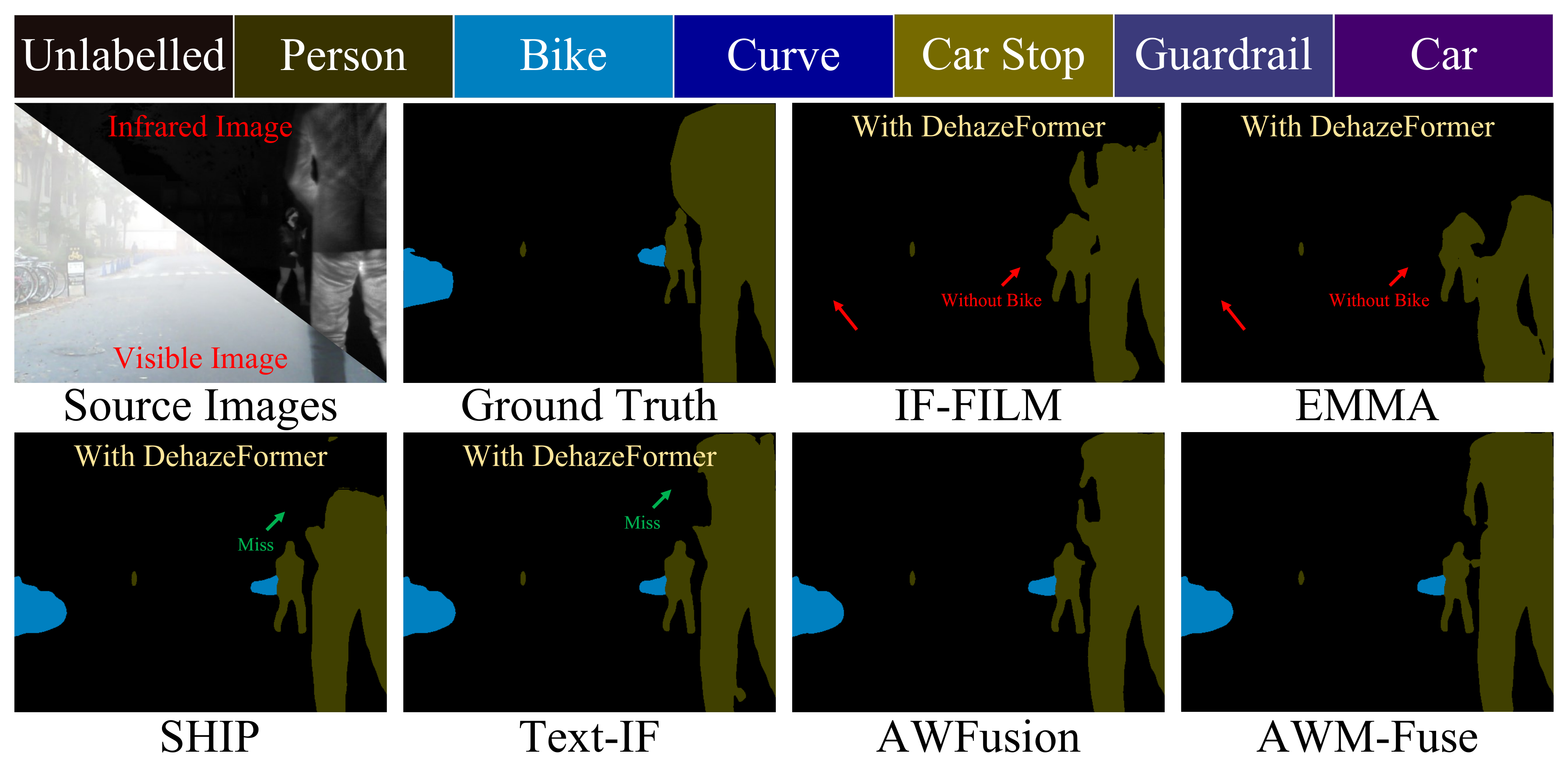}
   \caption{Segmentation comparisons of different methods under hazy conditions.}
   \label{fig6}
\end{figure}

\subsection{Image Fusion in Adverse Weather}
\noindent \textbf{Qualitative Comparison.}
As shown in Figure~\ref{fig3} to  Figure~\ref{fig5}, we compare the proposed algorithm with all comparison methods across three adverse weather scenes. The left part of each figure primarily highlights the interference removal performance of the proposed method. In the right part of each figure, the first row displays fusion results of the comparison methods in ideal conditions, the second row shows fusion with a baseline image restoration method, and the third row presents fusion results from an integrated model designed for adverse weather. We also evaluate our algorithm on clean source images, referred to as AWF-Fuse (In clean), to demonstrate its performance under ideal conditions. Difference maps are provided to illustrate variations between fusion results obtained with clean images and under adverse weather.

The comparison indicates that, under ideal conditions, all fusion methods capture features from different modalities. Although preprocessing degraded visible images can reduce interference from degraded data, this approach often overrelies on the initial recovery, as shown in Figure~\ref{fig3} to Figure~\ref{fig5}, where unresolved degradation affects fusion. AWFusion mitigates adverse weather effects, but residual artifacts from color distortion and incomplete degradation removal persist. In contrast, AWM-Fuse excels in degradation removal, color fidelity, and multi-modality information extraction. The difference maps confirm that our algorithm preserves most salient scene features. Finally, fusion results in clean images show that, despite training under adverse weather, our algorithm effectively learns multi-modality feature extraction, maintaining high-quality fusion in ideal conditions.

\begin{figure}[t]
  \centering
   \includegraphics[width=1\linewidth]{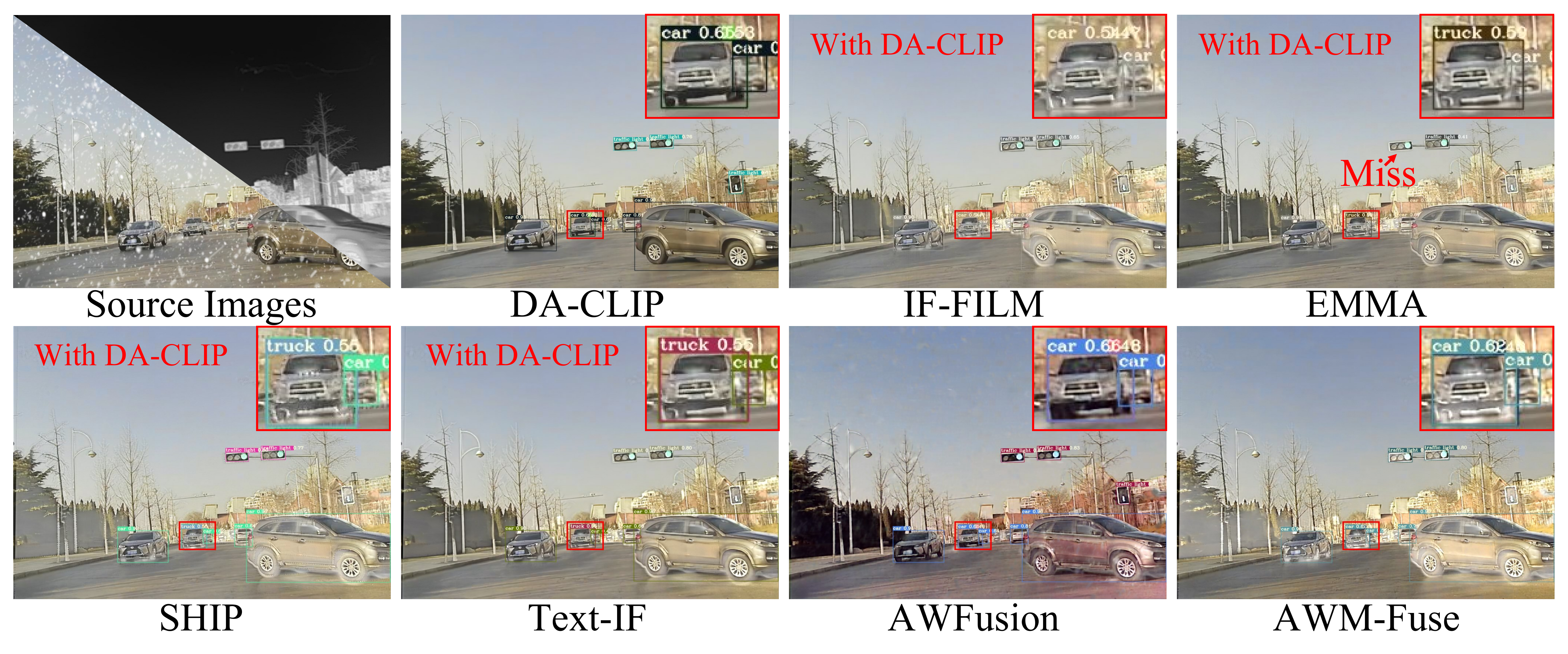}
   \caption{Detection comparisons of different methods under snow conditions.}
   \label{fig7}
\end{figure}

\noindent \textbf{Quantitative Comparison.}
Table~\ref{tab1} presents the quantitative comparison results of different methods under three types of adverse weather degradation. Our method ranks within the top two across all seven metrics, demonstrating that the proposed algorithm achieves superior performance in addressing multi-modality image fusion challenges under adverse weather conditions. In addition, for the integrated model, we observe that AWFusion performs well in both rain and snow scenes. However, the proposed method outperforms AWFusion in both image quality assessment metric scores and color recovery.

\subsection{Experiments on the Standard Dataset}
Although the proposed algorithm is designed for adverse weather, in order to verify its generalization ability in ideal scenarios, we conducted fusion experiments on three standard datasets, namely M3FD \cite{r17}, MSRS \cite{r62} and LLVIP \cite{r76}. As shown in Table~\ref{tab6}, the proposed method ranks in the top two in all datasets for at least six metrics, and its overall performance outperforms all the compared methods.

\subsection{Applications in Downstream Tasks}
We tested all methods on semantic segmentation and object detection to evaluate fused image quality. We used BANet \cite{r72} as the segmentation network and YOLOv7 \cite{r6} as the detector. Table~\ref{tab2} and Table~\ref{tab3} present quantitative comparisons of various methods on these tasks. As shown in Table~\ref{tab2}, our method achieves the highest intersection-over-union (IoU) scores across all categories, as well as the overall mean IoU in semantic segmentation. Furthermore, Table~\ref{tab3} shows that the proposed method yield the highest detection accuracy.

In addition, we present the qualitative comparison results. As shown in Figure~\ref{fig6}, the semantic information in the source image is diminished due to haze interference. While an image dehazing algorithm (DehazeFormer) was added as a pre-processing step for visible images in the first four comparison methods, they still exhibit segmentation errors with missing regions. Figure~\ref{fig7} shows the object detection comparison results. Although the snow remove method (DA-CLIP) was added to some fusion algorithms, the results failed to retain significant details due to the lack of synchronization between restoration and fusion. In contrast, the proposed algorithm effectively preserves scene information and outperforms others in object detection.

\begin{table*}[t]
\caption{Comparing quantitative results of different methods in object detection task. mAP denotes mean Average Precision.  The best scores are in bold, while the second-best scores are in blue.}
\begin{adjustbox}{width=\textwidth}
\begin{tabular}{c|c|c|c|cccccccc}
\hline
Source                  & Methods  & Pub.                & Baseline                  & People                       & Car                          & Bus                          & Lamp                         & Motorcycle                   & Truck                        & mAP@0.5                       & mAP@[0.5:0.95] \\ \hline
                        & TarDal \cite{r17}   & \textit{CVPR 2022}  & ×                         & 0.792                        & 0.877                        & 0.857                        & 0.585                        & 0.62                         & 0.713                        & 0.741                        & 0.462                                        \\
                        & LRRNet \cite{r32}   & \textit{TPAMI 2023} & ×                         & 0.785                        & {\color[HTML]{0000FF} 0.909} & 0.912                        & 0.764                        & {\color[HTML]{0000FF} 0.721} & {\color[HTML]{0000FF} 0.808} & 0.817                        & 0.484                                        \\
                        & TIM \cite{r79}      & \textit{TPAMI 2024} & ×                         & 0.731                        & 0.898                        & 0.878                        & 0.657                        & 0.656                        & 0.769                        & 0.765                        & 0.48                                         \\
\multirow{-4}{*}{Clean} & DRMF \cite{r89}     & \textit{MM 2024}    & ×                         & 0.699                        & 0.897                        & 0.896                        & {\color[HTML]{0000FF} 0.787} & 0.697                        & 0.784                        & 0.793                        & 0.494                                        \\ \hline
                        & FILM \cite{r80}  & \textit{ICML 2024}  &                           & 0.789                        & 0.891                        & 0.888                        & 0.678                        & 0.657                        & 0.743                        & 0.774                        & 0.483                                        \\
                        & EMMA \cite{r77}     & \textit{CVPR 24}    &                           & 0.785                        & 0.894                        & 0.901                        & 0.674                        & 0.68                         & 0.766                        & 0.783                        & 0.484                                        \\
                        & SHIP \cite{r88}     & \textit{CVPR 24}    &                           & 0.794                        & 0.9                          & 0.899                        & 0.754                        & 0.691                        & 0.793                        & 0.805                        & 0.499                                        \\
                        & Text-IF \cite{r65}  & \textit{CVPR 24}    & \multirow{-4}{*}{DA-CLIP} & {\color[HTML]{0000FF} 0.819} & 0.907                        & \textbf{0.921}               & 0.771                        & \textbf{0.723}               & 0.803                        & {\color[HTML]{0000FF} 0.824} & 0.522                                        \\ \cline{2-12} 
                        & AWFusion \cite{r91} & \textit{Arxiv24}    & ×                         & 0.791                        & 0.908                        & 0.903                        & 0.781                        & 0.714                        & 0.801                        & 0.816                        & {\color[HTML]{0000FF} 0.522}                 \\
\multirow{-6}{*}{Snow}  & AWM-Fuse & -                   & ×                         & \textbf{0.828}               & \textbf{0.913}               & {\color[HTML]{0000FF} 0.915} & \textbf{0.791}               & 0.717                        & \textbf{0.811}               & \textbf{0.829}               & \textbf{0.527}                               \\ \hline
\end{tabular}
\end{adjustbox}
\label{tab3}
\end{table*}

\subsection{Ablation Experiment}

\subsubsection{Effectiveness Analysis of Text} 

To further investigate the influence of textual information on network performance, we conduct three ablation experiments:  

\begin{itemize}
    \item \textbf{Noisy Text}: guiding the VLM model to generate descriptions that are deliberately inconsistent with the actual image content;  
    \item \textbf{Text-Reduced}: reducing the amount of descriptive text input by half;  
    \item \textbf{Text-Augmented}: introducing excessive and redundant textual descriptions.  
\end{itemize}

From the experiment results in Table~\ref{tab4}, we observed the following:
\textbf{(1)} Training with misguided textual descriptions leads to the most severe degradation in fusion performance;  
\textbf{(2)} Reducing textual input causes image feature--based metrics (e.g., $Q_{S}$ and $Q_{M}$) to drop significantly, suggesting that fewer textual cues weaken the ability of network to capture fine details;  
\textbf{(3)} Adding redundant text improves $Q_{S}$ and $Q_{M}$, but decreases perception-based metrics such as $Q_{CV}$ and $VIF$, implying that redundancy introduces deviations from natural human visual perception.

\begin{figure}[t]
  \centering
   \includegraphics[width=1\linewidth]{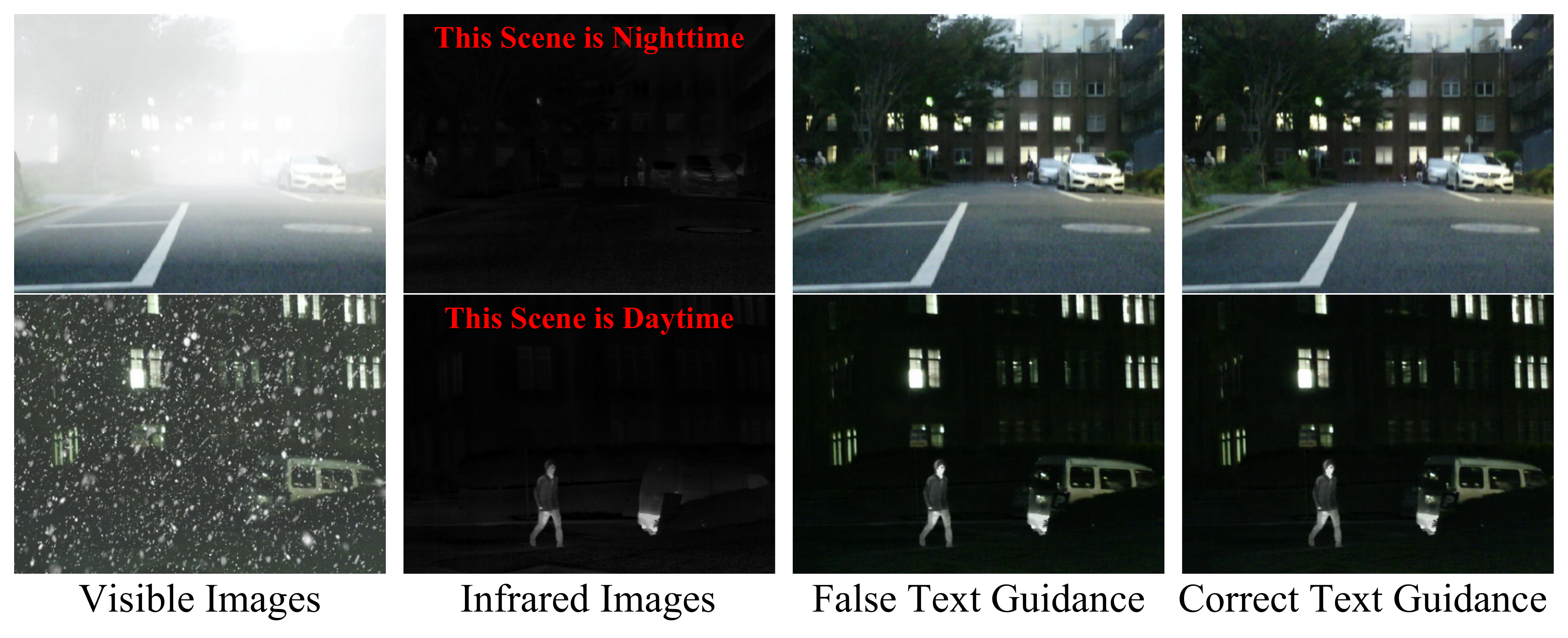}
   \caption{Fusion Results with Misleading Temporal Text.}
   \label{time}
\end{figure}

Since infrared images inherently lack temporal information (e.g., day or night), we further examine the robustness of textual guidance in this regard. We artificially inject erroneous temporal cues during fusion (e.g., labeling a nighttime scene as daytime). As shown in Figure~\ref{time}, our method still produces reasonable fusion results despite the biased cues. This demonstrates that text serves primarily as auxiliary guidance, while the main features are extracted from the original image, confirming the robustness.

\subsubsection{Effectiveness analysis of GTPM and LTPM} 
The GTPM module receives Caption data and text features encoded by CLIP to guide the network in identifying various degradation features. As shown in Table~\ref{tab4}, ablating the GTPM leads to declines in all metrics. Notably, the $VIF$ metric shows the largest drop, highlighting the difficulty of algorithm in balancing detail recovery with generalization, resulting in fusion outputs that retain some residual degradation.
The primary function of the LTPM module is to leverage detailed textual information, enabling the network to recover as much fine detail as possible from scenes obscured by degraded pixels. As shown in Table~\ref{tab4}, $SSIM$ is most affected when the LTPM is ablated, directly indicating a substantial loss of scene detail.

\subsubsection{Effectiveness analysis of Text (ChatGPT)}
We replaced all the detailed text generated by ChatGPT with Caption data as input to the LTPM. As shown in Table~\ref{tab4}, this replacement results in reduced scores for image feature-based metrics $Q_G$ and $Q_M$. However, since some text input is still retained, these metrics remain higher than those obtained by completely ablating the LTPM.

\subsubsection{Effectiveness analysis of VLM-driven Loss}
During training, text data generated by the VLM was used to supervise fusion result generation, ensuring alignment between image and text semantic information. The ablation results in Table~\ref{tab4} show a degree of performance degradation when this supervision is removed, highlighting the effectiveness of the proposed VLM-driven loss in achieving text-image feature alignment.

\begin{table*}[t]
\caption{Quantitative Comparative Structures in Ablation Studies. The best scores are in bold.}
\centering
\footnotesize
\begin{tabular}{ccccccccc}
\hline
\multicolumn{1}{c|}{Sources}                & \multicolumn{1}{c|}{Methods}                          & $Q_{G}$$\uparrow$                       & $Q_{M}$$\uparrow$                       & $Q_{S}$$\uparrow$                       & $Q_{CV}$$\downarrow$                      & $VIF$$\uparrow$                         & $SSIM$$\uparrow$                        & $SCD$$\uparrow$                         \\ \hline
\multicolumn{1}{c|}{}                       & \multicolumn{1}{c|}{Noise Text}                       & 0.4381                                  & 0.6053                                  & 0.8315                                  & 255.5851                                  & 0.3322                                  & 0.4260                                  & 1.5986                                  \\
\multicolumn{1}{c|}{}                       & \multicolumn{1}{c|}{Text-Reduced}                     & 0.4312                                  & 0.6761                                  & 0.8416                                  & 231.2314                                  & 0.3410                                  & 0.4332                                  & 1.6054                                  \\
\multicolumn{1}{c|}{}                       & \multicolumn{1}{c|}{Text-Augmented}                   & \textbf{0.4518}                         & \textbf{0.7105}                         & 0.8432                                  & 225.1728                                  & 0.3564                                  & 0.4345                                  & 1.6127                                  \\
\multicolumn{1}{c|}{\multirow{-4}{*}{Snow}} & \multicolumn{1}{c|}{\cellcolor[HTML]{EFEFEF}AWM-Fuse} & \cellcolor[HTML]{EFEFEF}0.4499          & \cellcolor[HTML]{EFEFEF}0.7046          & \cellcolor[HTML]{EFEFEF}\textbf{0.8456} & \cellcolor[HTML]{EFEFEF}\textbf{217.7734} & \cellcolor[HTML]{EFEFEF}\textbf{0.3573} & \cellcolor[HTML]{EFEFEF}\textbf{0.4406} & \cellcolor[HTML]{EFEFEF}\textbf{1.6263} \\ \hline
\multicolumn{1}{l}{}                        & \multicolumn{1}{l}{}                                  & \multicolumn{1}{l}{}                    & \multicolumn{1}{l}{}                    & \multicolumn{1}{l}{}                    & \multicolumn{1}{l}{}                      & \multicolumn{1}{l}{}                    & \multicolumn{1}{l}{}                    & \multicolumn{1}{l}{}                    \\ \hline
\multicolumn{1}{c|}{}                       & \multicolumn{1}{c|}{w/o GTPM}                         & 0.3537                                  & 0.6714                                  & 0.7481                                  & 235.3642                                  & 0.2812                                  & 0.3825                                  & 1.1227                                  \\
\multicolumn{1}{c|}{}                       & \multicolumn{1}{c|}{w/o LTPM}                         & 0.3298                                  & 0.5991                                  & 0.6827                                  & 227.2447                                  & 0.3145                                  & 0.3989                                  & 1.1892                                  \\
\multicolumn{1}{c|}{}                       & \multicolumn{1}{c|}{w/o Text (ChatGPT)}                & 0.3477                                  & 0.6734                                  & 0.7395                                  & 220.3748                                  & 0.3285                                  & 0.4137                                  & 1.2328                                  \\
\multicolumn{1}{c|}{}                       & \multicolumn{1}{c|}{w/o VLM-driven Loss}              & 0.4477                                  & 0.6892                                  & 0.6992                                  & 192.1521                                  & 0.3331                                  & 0.4305                                  & 1.2358                                  \\
\multicolumn{1}{c|}{\multirow{-5}{*}{Haze}} & \multicolumn{1}{c|}{\cellcolor[HTML]{EFEFEF}AWM-Fuse} & \cellcolor[HTML]{EFEFEF}\textbf{0.5297} & \cellcolor[HTML]{EFEFEF}\textbf{0.8193} & \cellcolor[HTML]{EFEFEF}\textbf{0.8392} & \cellcolor[HTML]{EFEFEF}\textbf{168.0220} & \cellcolor[HTML]{EFEFEF}\textbf{0.3936} & \cellcolor[HTML]{EFEFEF}\textbf{0.4631} & \cellcolor[HTML]{EFEFEF}\textbf{1.3416} \\ \hline
\end{tabular}
\label{tab4}
\end{table*}

\subsection{Extended Experiments in Real-World Scenarios}
\begin{figure}[t]
  \centering
   \includegraphics[width=1\linewidth]{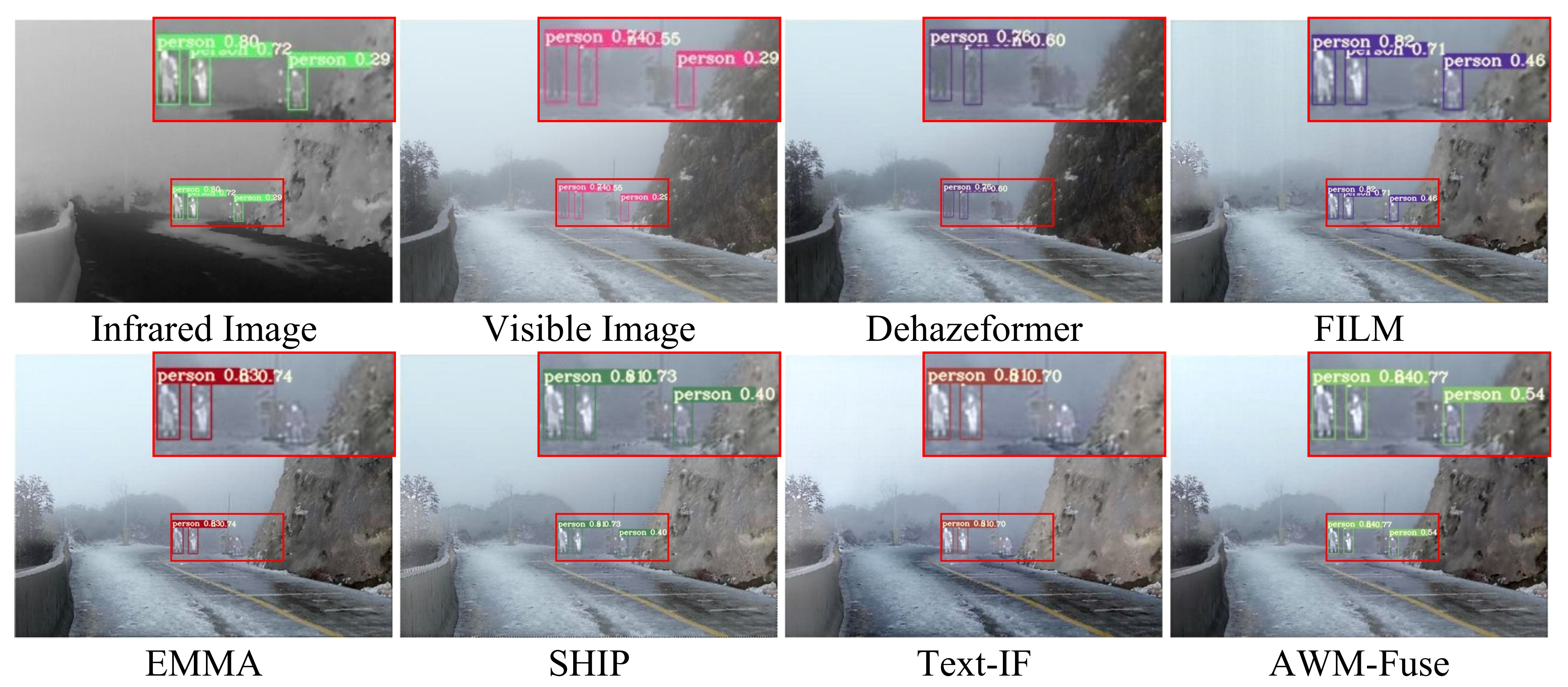}
   \caption{Fusion results of different algorithms in real-world scenarios.}
   \label{real}
\end{figure}

We conducted fusion experiments on real haze images in AWMM-100k. Due to the lack of available Ground Truth in real scenes for quantitative evaluation, this part of the experiment is analyzed in a qualitative manner for comparison. we introduce the YOLOv7 detector to identify the targets in the fusion results. The experimental results in Figure~\ref{real} show that even under real bad weather conditions, the proposed method can still effectively retain significant target information and obtain the highest detection accuracy on all target categories.

\subsection{Limitation}
Extracting features using the VLM model inevitably incurs additional computational overhead. Our proposed algorithm, when tested on images of $480 \times 640$ size, has FLOPs of $1145.5 G$ and parameters of $137.29 M$. In comparison, Text-IF, which incorporates the VLM model, has FLOPs of $1518.88 G$ and parameters of $89.01 M$. We recognize this as a limitation of the current work and plan to optimize computational efficiency in future iterations. However, despite the fact that the computational cost is not dominant, our method exhibits excellent fusion performance, outperforming many state-of-the-art image restoration and fusion models in our experiments.


\section{Conclusion}

In this paper, we propose a multi-modality image fusion method tailored for adverse weather. We leverage the image-to-text generation capability of ChatGPT to create text descriptions for source images under adverse weather and use BLIP to generate Captions. To  utilize this text, we design global and local textual feature perception modules. We also proposed a VLM-driven loss to align images and text features within the CLIP feature space. Finally, we validated the effectiveness of our method by extensive experiments across adverse weather conditions, clean scenarios, and downstream tasks.

\bibliographystyle{IEEEtran}
\bibliography{main}

\end{document}